\documentclass[sigconf]{aamas} 

\usepackage{balance}

\usepackage[inline]{enumitem}

\makeatletter
\@namedef{ver@lineno.sty}{9999/12/31}
\@namedef{opt@lineno.sty}{}
\makeatother
\usepackage[frozencache,cachedir=.]{minted}

\usepackage{caption}
\usepackage{subcaption}

\usepackage[subpreambles=true]{standalone}

\usepackage{xcolor}

\usepackage{mystyle}

\usepackage{orcidlink}

\setcopyright{ifaamas}
\acmConference[AAMAS '24]{Proc.\@ of the 23rd International Conference
on Autonomous Agents and Multiagent Systems (AAMAS 2024)}{May 6 -- 10, 2024}
{Auckland, New Zealand}{N.~Alechina, V.~Dignum, M.~Dastani, J.S.~Sichman (eds.)}
\copyrightyear{2024}
\acmYear{2024}
\acmDOI{}
\acmPrice{}
\acmISBN{}

\acmSubmissionID{1249}

\title{\papertitle{}: \papersubtitle{}}
\subtitle{Demonstration Track}

\author{Nathan Gavenski \orcidlink{0000-0003-0578-3086}}
\affiliation{
    \institution{King's College London}
    \city{London}
    \country{United Kingdom}}
\email{nathan.schneider_gavenski@kcl.ac.uk}
\orcid{0000-0003-0578-3086}

\author{Michael Luck \orcidlink{0000-0002-0926-2061}}
\affiliation{
  \institution{University of Sussex}
  \city{Sussex}
  \country{United Kingdom}}
\email{michael.luck@sussex.ac.uk}
\orcid{0000-0002-0926-2061}

\author{Odinaldo Rodrigues \orcidlink{0000-0001-7823-1034}}
\affiliation{
  \institution{King's College London}
  \city{London}
  \country{United Kingdom}}
\email{odinaldo.rodrigues@kcl.ac.uk}
\orcid{0000-0001-7823-1034}

\begin{abstract}
    Imitation learning field requires expert data to train agents in a task.
    Most often, this learning approach suffers from the absence of available data, which results in techniques being tested on its dataset.
    Creating datasets is a cumbersome process requiring researchers to train expert agents from scratch, record their interactions and test each benchmark method with newly created data.
    Moreover, creating new datasets for each new technique results in a lack of consistency in the evaluation process since each dataset can drastically vary in state and action distribution.
    In response, this work aims to address these issues by creating {\em Imitation Learning Datasets}, a toolkit that allows for:
    \begin{enumerate*}[label=(\roman*)]
        \item curated expert policies 
        with multithreaded support for faster dataset creation;
        \item readily available datasets and techniques with precise measurements; and 
        \item sharing implementations of common imitation learning techniques. 
    \end{enumerate*}
    \textbf{Demonstration link:} \url{https://nathangavenski.github.io/\#/il-datasets-video}
\end{abstract}


\newcommand{\BibTeX}{\rm B\kern-.05em{\sc i\kern-.025em b}\kern-.08em\TeX}

\makeatletter
\gdef\@copyrightpermission{
	\begin{minipage}{0.3\columnwidth}
		\href{https://creativecommons.org/licenses/by/4.0/}{\includegraphics[width=0.90\textwidth]{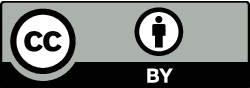}}
	\end{minipage}\hfill
	\begin{minipage}{0.7\columnwidth}
		\href{https://creativecommons.org/licenses/by/4.0/}{This work is licensed under a Creative Commons Attribution International 4.0 License.}
	\end{minipage}
	\vspace{5pt}
}
\makeatother

\begin{document}


\pagestyle{fancy}
\fancyhead{}


\maketitle 

\section{Introduction} \label{sec:introduction}
Creating imitation learning (IL) techniques requires researchers to gather \textit{expert} samples to train an agent in the desired task.
This process can be arduous since collecting expert samples usually involves either recording humans performing the task or training a new agent from scratch using another learning paradigm.
Moreover, creating new datasets for each new technique does not allow for evaluation consistency across different IL approaches~\cite{zheng2022taxonomies}.

When creating IL datasets, we must also consider which experts are used to collect data.
Most often, when a new expert is created, no information is provided about the quality of the data collected~\cite{belkhale2023data}, which can drastically affect the performance across different IL approaches~\cite{zheng2022taxonomies}.
Moreover, researchers must find available code and run it in any newly created dataset for consistent comparison.

Evaluating IL techniques in new datasets is time-consuming since:
\begin{enumerate*}[label=(\roman*)]
    \item not all techniques have code readily available;
    \item implementations might contain bugs; and 
    \item published versions might not support the environment the user is experimenting with. 
\end{enumerate*}
Testbeds, such as Gym~\cite{zenodo2023gymnasium}, help researchers to overcome problems with different environment support.
However, it is up to researchers to create code that supports all available environments, such as those with continuous and discrete actions and states spaces.

\begin{figure}[b]
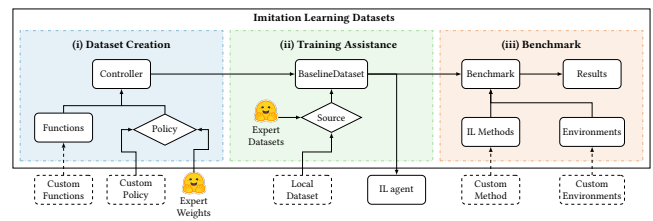

    \centering
    \includestandalone[width=\columnwidth]{figures/diagram}
    \caption{
        Visualisation of typical pipeline implementation.
    }
    \label{fig:il_process}
\end{figure}

In light of these issues, we have developed {\em Imitation Learning Datasets} (IL-Datasets)~\cite{ildatasets2023}, a toolkit that aims to help researchers through:
\begin{enumerate*}[label=(\roman*)]
    \item creating new datasets by allowing for faster \textit{multithreaded} creation and curated expert policies; \label{enum:creation}
    \item assisting the training of IL agents by sharing \textit{readily available datasets} with easily customisable data; and \label{enum:dataset} 
    \item benchmarking by \textit{providing results} for IL techniques in a diverse set of environments. \label{enum:benchamrk}
\end{enumerate*}
Figure~\ref{fig:il_process} shows each stage in a typical pipeline implementation.
In Step~\ref{enum:creation} users can use a `Controller' class that allows them to create datasets using \textit{expert policies} hosted on HuggingFace~\cite{huggingface2023} or \textit{custom-made} ones following the `Policy' interface.
With the dataset selected (based on the data provided on HuggingFace or locally), the user can instantiate the `BaselineDataset' to create training and evaluation datasets to train an agent in step~\ref{enum:dataset}.
Lastly, in step~\ref{enum:benchamrk}, users can specify a benchmarking strategy, including data and hyperparameters, and the toolkit will ensure that no \textit{leakage} (which refers to training data being present during test) will occur, recording the benchmarking details for reproducibility and better comparison with other work.

\section{Dataset Creation} \label{sec:creation}

The multithreaded `Controller' class allows users to execute functions that record the `Policy' experiences asynchronously.
This is a lightweight class since it creates a thread pool with a fixed number of threads (informed by the user) and spawns objects that will be executed for each episode instead of processes waiting for available resources.
Therefore, as the execution of each episode ends, the execution of a new episode starts ensuring nearly $100\%$ uptime in all threads.
IL-Datasets is agnostic to environment implementation (e.g., vectorized environments) since threads do not share memory pointers.
IL-Datasets allows users for the creation of new datasets using already curated policies ensuring for lower behaviour divergence across different datasets~\cite{belkhale2023data}.
In addition, users can also define custom-made policies to create new datasets while still benefiting from IL-Datasets features, and make them available for other users.

To create a new dataset a user simply needs to provide an `enjoy' function, which uses a policy to interact with the environment and collect samples during an episode; and a `collate' function that creates a single dataset file from all the recordings of the `enjoy' function. 
IL-Datasets provides functions for converting the newly created dataset into a HuggingFace dataset, which can later be updated to the platform if the user desires.
The following code snippet illustrates the creation of a new dataset using IL-Datasets.

\begin{minted}{python}
    from imitation_datasets.controller import Controller
    from imitation_datasets.functions import baseline_enjoy |\label{code:enjoy}|
    from imitation_datasets.functions import baseline_collate |\label{code:collate}|
    Controller( |\label{code:controler_start}|
        baseline_enjoy, baseline_collate, episodes=1000, threads=4 |\label{code:controller}|
    ).start({"game": "walker", "mode": "all"} |\label{code:controler_end}|
\end{minted}

\noindent
The `Controller' class (Lines~\ref{code:controler_start}-\ref{code:controler_end}) uses the already provided `enjoy' and `collate' functions, Lines~\ref{code:enjoy} and~\ref{code:collate}, respectively.
Therefore, creating a new dataset only requires $6$ lines of code (without losing the asynchronous multithread benefit).
We provide `enjoy' and `collate' functions (based on the StableBaselines~\cite{stablebaselines2018} pattern) for fast prototyping of new agents, where these functions will create a dataset containing `state', `action', `reward', `accumulated episode reward', and `episode start' (signalling which states are the first in each episode).  
Line~\ref{code:controler_end} starts the asynchronous multithread process for the register entry `walker', which has a curated expert for the `Walker2d-v3' environment.
IL-Datasets provides a list of registered environments with expert policies.\footnote{\url{http://github.com/NathanGavenski/IL-Datasets}}
Lastly, we allow customisable `enjoy' and `collate' functions to avoid the pitfalls from Imitations~\cite{gleave2022imitation} and StableBaselines~\cite{stablebaselines2018}, which have dataset creation functions but only support a strict format not suitable for most common state-of-the-art code available.
\balance
\section{Training Assistance} \label{sec:dataset}

IL-Datasets provides a `BaselineDataset' class that allows researchers to use custom-made~(Line~\ref{code:local}) or hosted data (Lines~\ref{code:hf}).
\begin{minted}{python}
    from src.imitation_datasets.dataset import BaselineDataset
    local = BaselineDataset("/path/to/local/file.npz") |\label{code:local}|
    hf = BaselineDataset("/path/to/hosted/data", source="hf" ) |\label{code:hf}|
\end{minted}

\noindent
It is important to note that even though these datasets are hosted on HuggingFace, once downloaded, the whole process can be executed offline if so needed.
The `BaselineDataset' class inherits the PyTorch Dataset class~\cite{pytorch2019} and returns a tuple of ($s_t$, $a_t$, $s_{t+1}$), where $s_t$ and $s_{t+1}$ are the current and next states, and $a_t$ is the action responsible for the state transition.
`BaselineDatasets' can also be inherited from other classes to support other formats, e.g., sequential data.
\footnote{\url{https://gist.github.com/NathanGavenski/ec904c7c3bf06b6361a0897b798206ac}}
By using IL-Datasets's\footnote{\url{https://nathangavenski.github.io/\#/IL-Datasets-data}} data, researchers can use up to $1,000$ episodes for each of the available environments.
These episodes can be divided between \textit{train}~(Line~\ref{code:train}) and \textit{evaluation}~(Line~\ref{code:eval}) splits:

\begin{minted}{python}
    from src.imitation_datasets.dataset import BaselineDataset
    dataset_train = BaselineDataset(..., n_episodes=100) |\label{code:train}|
    dataset_eval = BaselineDataset(..., n_episodes=100, split="eval")|\label{code:eval}|
\end{minted}

\noindent
In Line~\ref{code:train}, `n\_episodes' denotes the number of training episodes, e.g., $[0, 100)$, and in Line~\ref{code:eval}, it refers to the evaluation set interval, e.g., $[100, 1,000)$.
Each dataset published contains the expert policies used during creation and provides the average reward in cases where the \textit{performance} metric (also provided by IL-Datasets) is desired.
Moreover, all of these features exist within IL-Dataset to ensure \textit{consistency} in IL experiments.
By using the same dataset and splits, researchers can be sure that all trajectories remain the same through different runs.
\section{Benchmarking} \label{sec:benchmark}

The final IL-Datasets feature is \textit{benchmarking}, with which we aim to implement and test different IL techniques based on the available datasets.
We publish all data to help researchers reduce the amount of work to create IL techniques and to reduce the entry barrier to new researchers, but users' benchmarks will only be published if they desire to do so.
The benchmark trains each technique with the available data for $100,000$ epochs and, afterwards, evaluates each one using a specific set of seeds guaranteeing reproducibility and consistency across multiple executions.
These seeds are selected to reduce \textit{data leakage} and to make sure that the first state is not present in the training datasets.
Each technique is evaluated by executing the best model (according to the original work selection criteria) in each environment, and the \textit{average episodic reward} and \textit{performance} metrics are displayed in a list with all benchmark results published in the IL-Datasets page.
Furthermore, when training these techniques, IL-Datasets also uses specific seeds to guarantee that the training results will be the same for each method across multiple executions.
We note that this is done outside of Gym~\cite{zenodo2023gymnasium} environments since they do not support random number generators anymore.
Therefore, training IL-Datasets implementations without using these seeds will not guarantee the same results.

\section{Conclusion} \label{sec:conclusion}

In this paper, we described Imitation Learning Datasets: a toolkit to help researchers implement, train and evaluate IL agents.
IL-Datasets can be used to reduce comparison efforts while increasing consistency between published work.
It achieves this by offering:
\begin{enumerate*}[label=(\roman*)]
    \item fast and lightweight dataset creation through asynchronous multi-thread processes with curated expert policies that allow for no prior expert training and low behaviour divergence between different creations;
    \item readily available datasets, for fast prototyping of new techniques letting users only worry about model implementation; and 
    \item benchmarking results for IL techniques.
\end{enumerate*}
We believe that IL-Datasets will help facilitate the integration of new researchers and improve consistency across different IL work.


\begin{acks}
This work was supported by UK Research and Innovation [grant number EP/S023356/1], in the UKRI Centre for Doctoral Training in Safe and Trusted Artificial Intelligence (\url{www.safeandtrustedai.org}).
\end{acks}



\vfill\eject
\bibliographystyle{ACM-Reference-Format} 
\balance
\bibliography{bibliography}


\begin{thebibliography}{8}


\ifx \showCODEN    \undefined \def \showCODEN     #1{\unskip}     \fi
\ifx \showDOI      \undefined \def \showDOI       #1{#1}\fi
\ifx \showISBNx    \undefined \def \showISBNx     #1{\unskip}     \fi
\ifx \showISBNxiii \undefined \def \showISBNxiii  #1{\unskip}     \fi
\ifx \showISSN     \undefined \def \showISSN      #1{\unskip}     \fi
\ifx \showLCCN     \undefined \def \showLCCN      #1{\unskip}     \fi
\ifx \shownote     \undefined \def \shownote      #1{#1}          \fi
\ifx \showarticletitle \undefined \def \showarticletitle #1{#1}   \fi
\ifx \showURL      \undefined \def \showURL       {\relax}        \fi
\providecommand\bibfield[2]{#2}
\providecommand\bibinfo[2]{#2}
\providecommand\natexlab[1]{#1}
\providecommand\showeprint[2][]{arXiv:#2}

\bibitem[\protect\citeauthoryear{Belkhale, Cui, and Sadigh}{Belkhale
  et~al\mbox{.}}{2023}]%
        {belkhale2023data}
\bibfield{author}{\bibinfo{person}{Suneel Belkhale}, \bibinfo{person}{Yuchen
  Cui}, {and} \bibinfo{person}{Dorsa Sadigh}.} \bibinfo{year}{2023}\natexlab{}.
\newblock \showarticletitle{Data Quality in Imitation Learning}.
\newblock \bibinfo{journal}{\emph{arXiv}} (\bibinfo{year}{2023}).
\newblock
\showeprint[arXiv]{2306.02437v1}


\bibitem[\protect\citeauthoryear{Face}{Face}{2023}]%
        {huggingface2023}
\bibfield{author}{\bibinfo{person}{Hugging Face}.}
  \bibinfo{year}{2023}\natexlab{}.
\newblock \bibinfo{title}{Hugging Face}.
\newblock \bibinfo{howpublished}{Web Page}.
\newblock
\urldef\tempurl%
\url{https://huggingface.co/}
\showURL{%
\tempurl}


\bibitem[\protect\citeauthoryear{Gavenski}{Gavenski}{2023}]%
        {ildatasets2023}
\bibfield{author}{\bibinfo{person}{Nathan Gavenski}.}
  \bibinfo{year}{2023}\natexlab{}.
\newblock \bibinfo{title}{Imitation Learning Datasets}.
\newblock \bibinfo{howpublished}{GitHub Repository}.
\newblock
\urldef\tempurl%
\url{https://github.com/NathanGavenski/IL-Datasets}
\showURL{%
\tempurl}


\bibitem[\protect\citeauthoryear{Gleave, Taufeeque, Rocamonde, Jenner, Wang,
  Toyer, Ernestus, Belrose, Emmons, and Russell}{Gleave et~al\mbox{.}}{2022}]%
        {gleave2022imitation}
\bibfield{author}{\bibinfo{person}{Adam Gleave}, \bibinfo{person}{Mohammad
  Taufeeque}, \bibinfo{person}{Juan Rocamonde}, \bibinfo{person}{Erik Jenner},
  \bibinfo{person}{Steven~H. Wang}, \bibinfo{person}{Sam Toyer},
  \bibinfo{person}{Maximilian Ernestus}, \bibinfo{person}{Nora Belrose},
  \bibinfo{person}{Scott Emmons}, {and} \bibinfo{person}{Stuart Russell}.}
  \bibinfo{year}{2022}\natexlab{}.
\newblock \bibinfo{title}{imitation: Clean Imitation Learning Implementations}.
\newblock \bibinfo{howpublished}{arXiv:2211.11972v1 [cs.LG]}.
\newblock
\showeprint[arxiv]{2211.11972}~[cs.LG]
\urldef\tempurl%
\url{https://arxiv.org/abs/2211.11972}
\showURL{%
\tempurl}


\bibitem[\protect\citeauthoryear{Hill, Raffin, Ernestus, Gleave, Kanervisto,
  Traore, dhariwal, hesse, klimov, nichol, plappert, radford, schulman, sidor,
  and wu}{Hill et~al\mbox{.}}{2018}]%
        {stablebaselines2018}
\bibfield{author}{\bibinfo{person}{Ashley Hill}, \bibinfo{person}{Antonin
  Raffin}, \bibinfo{person}{Maximilian Ernestus}, \bibinfo{person}{Adam
  Gleave}, \bibinfo{person}{Anssi Kanervisto}, \bibinfo{person}{Rene Traore},
  \bibinfo{person}{prafulla dhariwal}, \bibinfo{person}{christopher hesse},
  \bibinfo{person}{oleg klimov}, \bibinfo{person}{alex nichol},
  \bibinfo{person}{matthias plappert}, \bibinfo{person}{alec radford},
  \bibinfo{person}{john schulman}, \bibinfo{person}{szymon sidor}, {and}
  \bibinfo{person}{yuhuai wu}.} \bibinfo{year}{2018}\natexlab{}.
\newblock \bibinfo{title}{Stable Baselines}.
\newblock
  \bibinfo{howpublished}{\url{https://github.com/hill-a/stable-baselines}}.
\newblock


\bibitem[\protect\citeauthoryear{Paszke, Gross, Massa, Lerer, Bradbury, Chanan,
  Killeen, Lin, Gimelshein, Antiga, Desmaison, Kopf, Yang, DeVito, Raison,
  Tejani, Chilamkurthy, Steiner, Fang, Bai, and Chintala}{Paszke
  et~al\mbox{.}}{2019}]%
        {pytorch2019}
\bibfield{author}{\bibinfo{person}{Adam Paszke}, \bibinfo{person}{Sam Gross},
  \bibinfo{person}{Francisco Massa}, \bibinfo{person}{Adam Lerer},
  \bibinfo{person}{James Bradbury}, \bibinfo{person}{Gregory Chanan},
  \bibinfo{person}{Trevor Killeen}, \bibinfo{person}{Zeming Lin},
  \bibinfo{person}{Natalia Gimelshein}, \bibinfo{person}{Luca Antiga},
  \bibinfo{person}{Alban Desmaison}, \bibinfo{person}{Andreas Kopf},
  \bibinfo{person}{Edward Yang}, \bibinfo{person}{Zachary DeVito},
  \bibinfo{person}{Martin Raison}, \bibinfo{person}{Alykhan Tejani},
  \bibinfo{person}{Sasank Chilamkurthy}, \bibinfo{person}{Benoit Steiner},
  \bibinfo{person}{Lu Fang}, \bibinfo{person}{Junjie Bai}, {and}
  \bibinfo{person}{Soumith Chintala}.} \bibinfo{year}{2019}\natexlab{}.
\newblock \showarticletitle{{PyTorch: An Imperative Style, High-Performance
  Deep Learning Library}}. In \bibinfo{booktitle}{\emph{Advances in Neural
  Information Processing Systems 32}},
  \bibfield{editor}{\bibinfo{person}{H.~Wallach},
  \bibinfo{person}{H.~Larochelle}, \bibinfo{person}{A.~Beygelzimer},
  \bibinfo{person}{F.~d'Alché Buc}, \bibinfo{person}{E.~Fox}, {and}
  \bibinfo{person}{R.~Garnett}} (Eds.). \bibinfo{publisher}{Curran Associates,
  Inc.}, \bibinfo{pages}{8024--8035}.
\newblock
\urldef\tempurl%
\url{http://papers.neurips.cc/paper/9015-pytorch-an-imperative-style-high-performance-deep-learning-library.pdf}
\showURL{%
\tempurl}


\bibitem[\protect\citeauthoryear{Towers, Terry, Kwiatkowski, Balis, de~Cola,
  Deleu, Goulão, Kallinteris, Kg, Krimmel, Perez-vicente, Pierré, Schulhoff,
  Tai, Shen, and Younis}{Towers et~al\mbox{.}}{2023}]%
        {zenodo2023gymnasium}
\bibfield{author}{\bibinfo{person}{Mark Towers}, \bibinfo{person}{Jordan~K
  Terry}, \bibinfo{person}{Ariel Kwiatkowski}, \bibinfo{person}{John~U. Balis},
  \bibinfo{person}{Gianluca de Cola}, \bibinfo{person}{Tristan Deleu},
  \bibinfo{person}{Manuel Goulão}, \bibinfo{person}{Andreas Kallinteris},
  \bibinfo{person}{Arjun Kg}, \bibinfo{person}{Markus Krimmel},
  \bibinfo{person}{Rodrigo Perez-vicente}, \bibinfo{person}{Andrea Pierré},
  \bibinfo{person}{Sander Schulhoff}, \bibinfo{person}{Jun~Jet Tai},
  \bibinfo{person}{Andrew Tan~Jin Shen}, {and} \bibinfo{person}{Omar~G.
  Younis}.} \bibinfo{year}{2023}\natexlab{}.
\newblock \bibinfo{title}{gymnasium}.
\newblock
\newblock
\urldef\tempurl%
\url{https://doi.org/10.5281/zenodo.8127026}
\showDOI{\tempurl}


\bibitem[\protect\citeauthoryear{Zheng, Verma, Zhou, Tsang, and Chen}{Zheng
  et~al\mbox{.}}{2022}]%
        {zheng2022taxonomies}
\bibfield{author}{\bibinfo{person}{Boyuan Zheng}, \bibinfo{person}{Sunny
  Verma}, \bibinfo{person}{Jianlong Zhou}, \bibinfo{person}{Ivor~W. Tsang},
  {and} \bibinfo{person}{Fang Chen}.} \bibinfo{year}{2022}\natexlab{}.
\newblock \showarticletitle{Imitation Learning: Progress, Taxonomies and
  Challenges}.
\newblock \bibinfo{journal}{\emph{{IEEE} Transactions on Neural Networks and
  Learning Systems}} (\bibinfo{year}{2022}), \bibinfo{pages}{1--16}.
\newblock
\urldef\tempurl%
\url{https://doi.org/10.1109/tnnls.2022.3213246}
\showDOI{\tempurl}


\end{thebibliography}


\end{document}